\documentclass[a4paper,fleqn]{cas-dc}
\usepackage{xcolor}

\usepackage[defaultcolor=red]{changes}

\usepackage[numbers]{natbib}

\def\tsc#1{\csdef{#1}{\textsc{\lowercase{#1}}\xspace}}
\tsc{WGM}
\tsc{QE}

\usepackage[defaultcolor=red]{changes} 
\usepackage{subfig}
\usepackage{xcolor}
\usepackage{soul}
\usepackage{algorithm}
\usepackage{algorithmic}
\definecolor{red1}{RGB}{255, 240, 240} 
\definecolor{red2}{RGB}{255, 220, 220} 
\definecolor{red3}{RGB}{255, 180, 180} 
\definecolor{red4}{RGB}{255, 140, 140} 
\definecolor{red5}{RGB}{255, 50, 50}

\begin{document}
\let\WriteBookmarks\relax
\def\floatpagepagefraction{1}
\def\textpagefraction{.001}

\shorttitle{DARF: Depth-Aware Generalizable Neural Radiance Field}   

\shortauthors{Y. Shi et al.}  

\newcommand{\yue}[1]{\textcolor{red}{\textbf{yue:} #1}}
\title [mode = title]{DARF: Depth-Aware Generalizable Neural Radiance Field}  

\author[1]{Yue Shi}
\author[1]{Dingyi Rong}
\author[1]{Chang Chen}
\author[1]{Chaofan Ma}
\author[1]{Bingbing Ni}
\cormark[1]
\cortext[1]{Corresponding author} 
\author[1]{Wenjun Zhang}

\affiliation[1]{organization={School of Electronics, Information and Electrical Engineering, Shanghai Jiao Tong University},
            city={Shanghai},
            postcode={200240}, 
            country={China}}


\begin{abstract}
Neural Radiance Field (NeRF) has revolutionized novel-view rendering tasks and achieved impressive results. However, the inefficient sampling and per-scene optimization hinder its wide applications. Though some generalizable NeRFs have been proposed, the rendering quality is unsatisfactory due to the lack of geometry and scene uniqueness. To address these issues, we propose the Depth-Aware Generalizable Neural Radiance Field (DARF) with a Depth-Aware Dynamic Sampling (DADS) strategy to perform efficient novel view rendering and unsupervised depth estimation on unseen scenes without per-scene optimization. Distinct from most existing generalizable NeRFs, our framework infers the unseen scenes on both pixel level and geometry level with only a few input images. By introducing a pre-trained depth estimation module to derive the depth prior, narrowing down the ray sampling interval to the proximity space of the estimated surface, and sampling in expectation maximum position, we preserve scene characteristics while learning common attributes for novel-view synthesis. Moreover, we introduce a Multi-level Semantic Consistency loss (MSC) to assist with more informative representation learning. Extensive experiments on indoor and outdoor datasets show that compared with state-of-the-art generalizable NeRF methods, DARF reduces samples by 50\%, while improving rendering quality and depth estimation. Our code is available on \href{https://github.com/shiyue001/GARF.git}{https://github.com/shiyue001/DARF.git}.
\end{abstract}

\maketitle



\section{Introduction} \label{introduction}
Novel-view synthesis is a vital research topic in computer vision, playing a key role in augmented reality, creative interaction, 3D reconstruction, etc. However, the most popular novel-view synthesis methods represented by NeRF \cite{mildenhall2020nerf} are still facing many challenges, especially the efficiency and generalizability of both rendering and geometry estimation. Therefore, designing a depth-aware generalizable neural radiance field is meaningful.

The efficiency of the neural radiance field mainly depends on its sampling rule and training strategy. For the sampling process, the original NeRF and many follow-up works \cite{meng2021gnerf,yu2021pixelnerf,wang2021ibrnet} query points along camera rays in the whole volume. A series of methods \cite{NSVFliu2020, PLENOCTRREE} combine radiance field with volumetric representation and prune redundant parts to speed up rendering, taking huge memory during training. More importantly, the consumption of time and computing resources in training cannot be ignored. Therefore, a scene-adaptive importance sampling scheme needs to be designed to speed up training and rendering. It is therefore plausible to make NeRF generalize to unseen scenes and generate novel-view images based on only a limited number of input views and images. Recent works explore using feature learning to make neural radiance field generalizable, \textit{e.g.}, PixelNeRF \cite{yu2021pixelnerf}, which takes spatial image features aligned to pixels. The introduction of features transforms NeRF from a single-scene optimization approach to a learning-based model that leverages input images and camera information, enabling improved generalization across diverse scenes. However, learning common features from multiple scenes inevitably compromises the uniqueness of the scene, thereby reducing rendering quality and depth estimation accuracy. \textbf{Hence, a framework for learning commonalities while preserving scene uniqueness is urgently needed.}

\begin{figure*}[ht]
\centering
\includegraphics[width=1\linewidth]{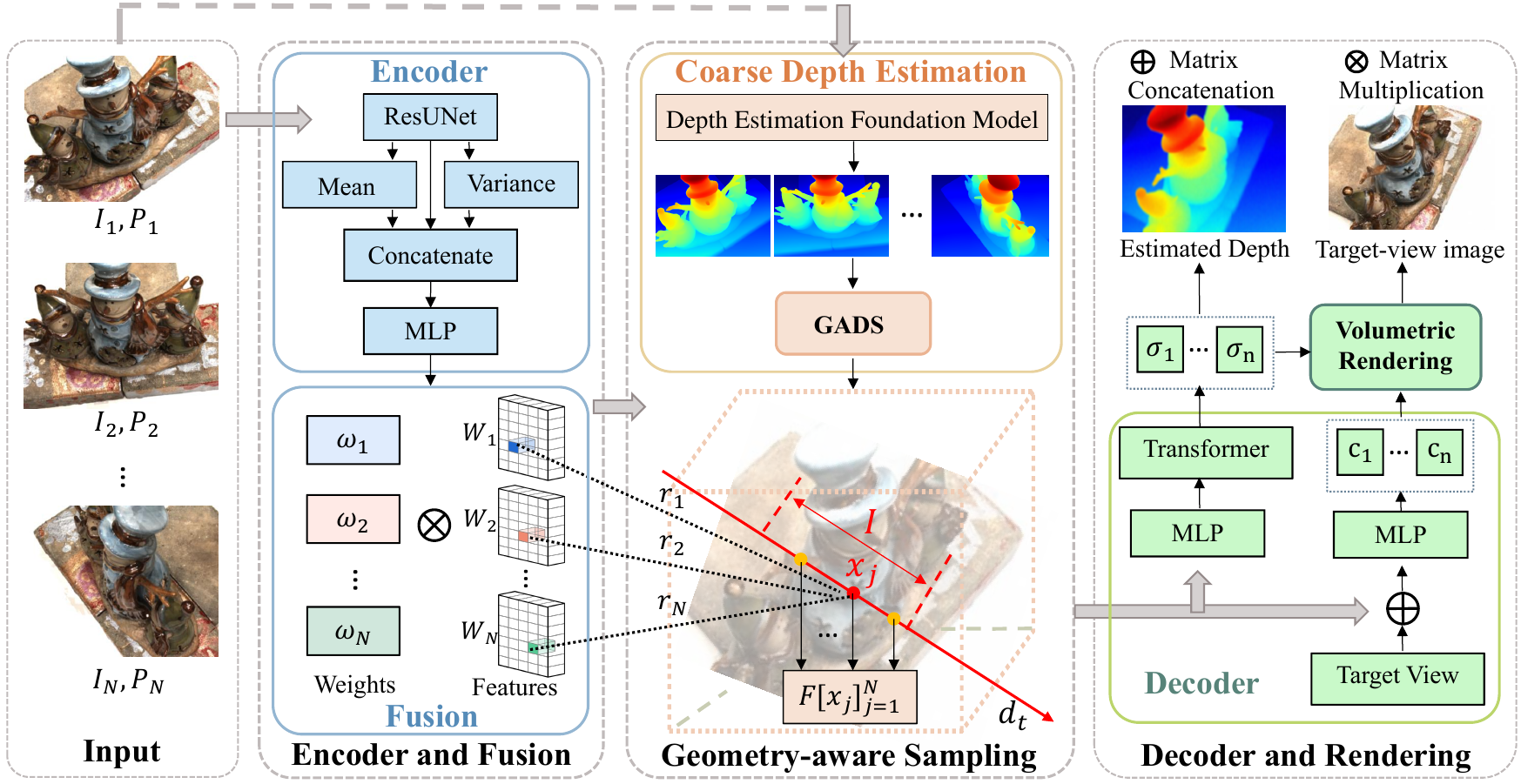}
\caption{Our framework. The proposed Depth-Aware Generalizable Neural Radiance Field (DARF) consists of three parts. First, deep convolutional features are extracted from input images to form learnable features of the scene. Second, a dynamic depth-aware sampling strategy is derived, based on the estimated depth prior provided by the pre-trained depth estimation foundation model. Finally, a decoder module is designed to predict color and density to render novel-view images along with fine depth map inference in a joint manner.}
\label{framework}
\end{figure*}

To address these issues, we propose a depth-aware generalizable radiance field (DARF) with a depth-aware dynamic sampling (DADS) strategy, which is capable of quickly recovering the unseen scene at both pixel-level and geometry-level, depending on only a few input images. The model learns common attributes of novel-view synthesis in different scenes to acquire generalization to unseen scenes and keeps specific to each scene based on the prior of its estimated depth. The proposed framework is shown in Fig. \ref{framework}, which contains a depth estimation module, a feature extraction and fusion module, a depth-aware dynamic sampling module, and a volumetric rendering module. First, we use a pre-trained depth estimation network to get depth maps for scenes. Based on the initial depth prior, we narrow the ray sampling interval down to the proximity space of the estimated surface. In the sampling interval, we propose a dynamic \emph{Predict-then-Refine} strategy to progressively approach the most likely location of the surface. For each sampling point, we derive its feature by fusing corresponding multi-view feature vectors learned by a feature extraction network with a set of learnable weights. Finally, we decode the feature of per point into density and color, which are used for volumetric rendering. In order to make the network learn more comprehensive representation, we introduce a Multi-level Semantic Consistency loss (MSC), which integrates both pixel-level loss and semantic feature matching loss to constrain the global similarity of predicted images and the ground truth with limited computational cost. Extensive experiments on indoor and outdoor datasets show that compared with existing generalizable NeRF methods, DARF reduces sampling points by 50\%, while improving rendering quality and depth estimation without per-scene optimization.

\section{Related Work}
\label{relatedwork}
\textbf{Novel View Synthesis.} 
Novel view synthesis is to generate the image of the target view from the inputted images.  
It has two main branches, including the weighted sum of reference images and rendering from 3D representation. The first line of work synthesizes novel views from a set of reference images by a weighted blending of reference pixels \cite{debevec1996modeling,gortler1996lumigraph,levoy1996light}. Blending weights are computed based on ray-space proximity \cite{levoy1996light} or approximate proxy geometry \cite{buehler2001unstructured,debevec1996modeling,heigl1999plenoptic}. 
In recent works, researchers have proposed improved methods for computing proxy geometry \cite{chaurasia2013depth,hedman2016scalable}, optical flow correction \cite{casas20154d,du2018montage4d,eisemann2008floating}, and soft blending \cite{penner2017soft,riegler2020free}. While these methods can handle sparser views than other approaches, they are fundamentally limited by the performance of 3D reconstruction algorithms and have difficulty in low-textured regions, reflective regions, or partially translucent surfaces.  

Another line of work mainly uses discrete volumetric representations. Prior to implicit representations, popular approaches for view synthesis used explicit representations like multi-plane images (MPIs)\cite{flynn2019deepview,flynn2016deepstereo,li2020crowdsampling,mildenhall2019local,srinivasan2019pushing,zhou2018stereo}, meshes \cite{hu2021worldsheet,shih20203d}, or point clouds \cite{wiles2020synsin} to render novel views. With the development of deep learning, the latest methods leverage convolutional neural networks (CNNs) to predict volumetric representations stored in voxel grids or multi-plane images. However, explicitly processing and storing voxel results in a large memory footprint, further restricting the resolution of the results.

\textbf{NeRF and Generalizable NeRF.} Neural radiance field (NeRF) \cite{mildenhall2020nerf} revolutionized the novel-view synthesis by introducing an implicit function, bringing out impressive results and various applications \cite{zhoupeng2023cips, meng2021gnerf, niemeyer2021giraffe, cao2023ciaosr, liu2024genn2n, Display_teeth, Display_InteractiveNeRF}. 
While NeRF and its follow-up works have achieved impressive results, they must be optimized for each new scene, requiring a large number of input images and taking days to converge. Though several methods have been proposed to train on fewer views or render faster \cite{PLENOCTRREE, nerfid}, the non-generalizability of the models still hinders the applications.

A pioneer generalizable NeRF framework is proposed by pixelNeRF \cite{yu2021pixelnerf}. By introducing image features, pixelNeRF can be trained across multiple scenes. To improve image quality when generalizable to novel scenes, IBRNet \cite{wang2021ibrnet} designs a ray transformer to take advantage of the contextual information on the ray. \cite{liu2024pixel} uses Pixel-Learnable 3D Lookup Table
(3DLUT) for real-time image enhancement, which provides the possibility of improving the novel-view rendering results.  
GRF \cite{trevithick2021grf} learns local features for each pixel in images and integrates an attention mechanism \cite{trevithick2021grf}. Stereo Radiance Fields (SRF) \cite{srf} introduces patch matching into neural view synthesis to realize sparse inputs. MVSNeRF \cite{chen2021mvsnerf} leverages plane-swept cost volumes to improve the performance of the generalizable NeRF. However, the above generalizable methods inherit the sampling strategy of the original NeRF \cite{mildenhall2020nerf}, which samples in the whole space inefficiently. Generalizable learning makes the network learn only the interpolation of images in the pixel scale, without geometry learning. The lack of geometric information leads to constraints on generalization capabilities, inefficient rendering, and blurring of novel-view synthesis, especially when the input views are sparse. Besides, recent depth estimation model \cite{depthany} learned on large amounts of data have achieved unprecedented results, which provide additional geometric priors for image data. For the use of depth, DS-NeRF \cite{deng2021depth} proposes a loss to take advantage of depth, exploiting the insight that sparse depth supervision can be used to regularize the learned geometry. But the constraint in \cite{deng2021depth} is sparse and the depth prior limits the prediction. We propose a DARF framework and DADS scheme to realize geometry-aware sampling and learning.

\section{Methodology}
The original NeRF \cite{mildenhall2020nerf} optimizes each scene independently, hindering real-time scene reconstruction and rendering application. Though some generalizable NeRF-based methods \cite{yu2021pixelnerf,wang2021ibrnet,chen2021mvsnerf} attempt to learn common attributes of novel-view synthesis in varying scenes, they are still far from achieving realistic rendering quality or handling large scene and view-angle changes, since these methods lack a principled way to encode sufficient inherent geometric information among multi-view images. Besides, sampling in an unconstrained high-dimensional space makes these NeRFs computationally expensive, hindering their potential application in portable platforms. Focusing on the above problems, we design a Depth-Aware Generalizable Neural Radiance Field (DARF) model to realize more efficient and higher-fidelity novel scene rendering and depth estimation based on depth priors. The overall framework is shown in Fig.~\ref{framework}, which consists of two novel components, \textit{i.e.}, a geometry representation learning module along with position and orientation encodings to achieve both generalization and scene-specific details, as well as a Depth-Aware Dynamic Sampling Strategy (DADS) to realize more accurate and efficient sampling via coarsely inferred depth (\textit{i.e.}, depth estimation) guidance. In addition, we propose a Multi-level Semantic Consistency Loss (MSC) to facilitate learning more informative representation.

\subsection{Generalizable Neural Radiance Field}
\label{section1}
The goal is to predict color and density at the point with coordinate $x$ in 3D space, given $ \mathcal{I} = \left\{ I_{i}, P_{i}\right\}_{i=1}^{N}$, a set of $N$ input reference images, with camera parameters. Neural radiance fields were initially proposed in NeRF \cite{mildenhall2020nerf} and become popular recently for their geometric superiority in processing multi-view images. The main idea of NeRF is utilizing an implicit deep model to fit the distribution of color and density of points, which are parameterized by real-valued coordinates and viewing directions of the considered 3D volume, supervised by images captured from multiple view angles. However, NeRF has no cross-scene generalization capability, due to the large semantic gap between input coordinates and output image pixel values, \textit{i.e.}, with no semantic features that can encode geometric commonalities. Therefore, in order to cope with unseen scenes, we design a geometric structure representation of learnable NeRF by a deep encoder-decoder framework. Specifically, we extract deep convolution features representing the geometry, shape, texture, etc. of input images by an encoder module and fuse multi-view features for representing each point in the 3D space semantically. Then, this novelly introduced scene representation combined with the original coordinate query eliminates the semantic gap of implicit mapping and enhances generalization via commonality learning. In this sense, our goal is to learn a function $f_{GARF}$ which maps the point $x_{j}$, feature $F(x_{j})$ and view direction $r$ of the point to color $c$ and opacity $\sigma$, described as follows:
\begin{equation}
   f_{GARF}:\left( F(x_{j}),x_{j},r \right)\mapsto \left( c,\sigma \right).
\end{equation}
In the following, we will introduce the proposed encoder-decoder structure in detail.

\textbf{Encoder and Feature Fusion.} In contrast to NeRF, where the inputs are point coordinates without scene-specific information, we condition our prediction on the reference images. Specifically, we first use a U-ResNet to extract the feature ${M_{i}\in\mathcal{R}^{H\times W\times z} }$ of the image $I_{i}\in \left [0,1\right]^{H\times W\times 3}$. The skip connection of U-ResNet better reveals low-level information in images while preserving global scene structure, which helps generalize to realistic novel-view images. In addition, to check the consistency among the features ${M_{i}}$, we concatenate the image feature with its mean and variance map. The concatenated feature is fed in a PointNet-like MLP to integrate local and global information and generate multi-view aware features $W_{n}$, along with a set of learnable weights $\omega_{n}$.
Then, we fuse the features of different $N$ views together. Previous feature fusing methods in multi-view synthesis use average weighting or weighting based on the distance between the reference view and target view. Actually, the correlation between each 3D point and the multi-view images is different, which cannot be described by the same weight. SRF \cite{srf} proposes using patch matching to set reasonable weights, depending on huge memory. We achieve feature fusion in point-wise, projecting every 3D point into each reference view, extracting the corresponding local features of each view, and fusing them together. We use a set of learnable weights for every point to fuse multi-view features. For a target view, a camera ray can be parameterized as $\mathbf{r}(t)= \mathbf{o} + t\mathbf{d}$, with the ray origin (camera center) $\mathbf{o}\in R^{3}$ and ray unit direction vector $\mathbf{d}$. Taking point $x_{j}$ in Fig.~\ref{framework} as an example, we get its feature by reflecting it to feature maps of $N$ input images along corresponding viewing direction $r_{i}$ as Eq.(\ref{r_{i}}). Consequently, we get the feature $V_{j}^{i}$ from view $i$ as Eq.(\ref{V_{j}^{i}}), where ${\rm proj}(\cdot)$ represents the coordinate obtained by the projection of point $x_{j}$ onto feature cube $W_{i}$ along the ray $r_{i}$. $\mathbf{x}_{j}$ is the coordinate of the point ${x}_{j}$.
\begin{equation}
\label{r_{i}}
    \mathbf{r}_{i}= \mathbf{o}_{i} + t\left ( \mathbf{x}_{j}- \mathbf{o}_{i}\right),
\end{equation}
\begin{equation}
\label{V_{j}^{i}}
    V_{j}^{i}=W_{i}\left[\ {\rm proj}(\mathbf{x}_{j}, \mathbf{r}_{i})\,\right].
\end{equation}
Because of the occlusion and illumination, different perspectives contribute differently to the same point. Therefore, to realize more accurate and view-dependent results, we utilize a set of learnable parameters $w_{i}$ to weighting $N$ different views. The fused feature $F(x_{j})$ for point $x_{j}$ is:
\begin{equation}
    F(x_{j})=\sum_{i=1}^{N}w_{i}V_{j}^{i}.
\end{equation}
\begin{figure*}[h]
\centering
\includegraphics[width=1\linewidth]{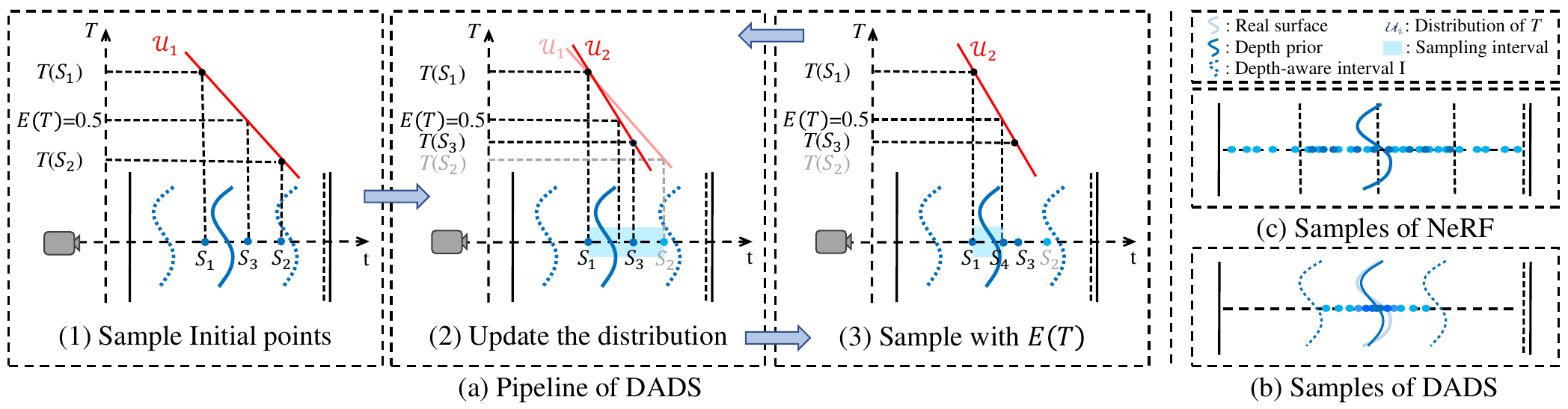}
\caption{Illustration of the depth-aware dynamic sampling. Compared with the sampling strategy in NeRF, our proposed DADS strategy tends to distribute more sample points near the surface.}
\label{sampling illu}
\end{figure*}
\textbf{Decoder and Volumetric Rendering.} We have already introduced how to represent a volumetric radiance field conditioned on deeply learned scene features. Here, we introduce how the decoder maps the feature representation into color and density for every sample point. In contrast to most NeRF models which utilize MLP as the decoding structure, in this work we adopt the transformer structure in \cite{wang2021ibrnet} to perform feature decoding of sample sequences on the viewing ray in a context-aware manner. Namely, spatially nearby samples present correlated characteristics; however, MLP independently processes each 3D point without considering the geometry correlations among adjacent samples. On the contrary, transformer-based structures perform self-attention operations, so that contextual dependencies are well modeled and all samples are inferred in a joint way. Based on this decoder structure, we predict the blending weights of the image colors for variation reduction to improve the visual quality of the synthesized results. After getting color and density values at continuous 5D locations, we use the volumetric rendering method \cite{volumetric1996, DVR_2020_CVPR, Vol_2020_CVPR} to calculate the color for every pixel. More concretely, we first query colors and densities of $M$ samples on the viewing ray, and then we derive the pixel color $\hat{C_{r}}$ along the ray $\mathbf{r}$ as: 
\begin{equation}
  \hat{C_{r}}=\sum_{j=1}^{M}T_{j}\left (1-{\rm  {\rm exp}}\left(-\sigma _{j}\delta_{j}\right) \right )c_{j}, 
\end{equation} 
\begin{equation}
  T_{j}=  {\rm exp}\left(-\sum_{k=1}^{j-1}\sigma _{k}\delta_{k}\right),
\end{equation}
where samples from $1$ to $M$ are sorted according to ascending depth values. $c_{j}$ and $\sigma_{j}$ denote the color and density of the $j$-th sample on the ray, respectively. $\delta_{j}$ is the distance between $j$-th sample and $(j{-}1)$-th sample. We show our innovation on how to accelerate this sampling process in the following.

\subsection{Depth-Aware Dynamic Sampling}
\label{DADS}
Sampling and integrating along the ray is the most computationally expensive stage in neural rendering. The vanilla sampling strategy in NeRF, which explores the entire 3D volume, not only wastes a lot of computation in empty space but also brings redundant points far away from the real surface, leading to blurry and wrong geometry. To this end, we propose a Depth-Aware Dynamic Sampling scheme (DADS) to realize more accurate and efficient sampling. We show the samples of DADS and NeRF in Fig.~\ref{sampling illu} (b) and (c). The main idea of DADS is two-fold: (\textit{i}) leveraging the depth prior estimated from input images to get a depth-aware sampling interval, and (\textit{ii}) a dynamic \emph{\emph{\emph{Predict-then-Refine}}} sampling strategy to adaptively sample points that are most likely to be close to the geometry surface. Note that DADS realizes both high efficiency and geometry accuracy.

\textbf{Depth-aware Sampling Interval.} To narrow down the sampling space, we estimate the geometry surface of the scene and sample in the nearby region. As the coarse depth estimation shown in Fig.~\ref{framework}, we use the pre-trained Depth Anything v2 \cite{depthany} to derive the depth map for input images. After rescaling the depth to the DARF space, we exploit it as a geometry prior for sampling.
Then, based on the coarse depth $d_{c}$, we restrict the sampling operation in an interval $D$ that encloses the space where the surface is most likely to appear, denoted by
\begin{equation}
\label{range}
  D= \left [ d_{c}-\Delta d, d_{c}+\Delta d \right ].
\end{equation}
In this way, we adapt the sampling process to scene geometries. Moreover, because of the adjustable interval $\Delta d$, inaccurate depth estimation has the opportunity to be corrected in the subsequent optimization of the neural radiance field. 

\textbf{Dynamic Sampling Strategy.} When the sampling interval is determined, we propose a dynamic sampling strategy (\textit{i.e.}, a new importance sampling scheme), which is shown in Fig.~\ref{sampling illu} (a). Compared with stratified sampling in \cite{mildenhall2020nerf,wang2021ibrnet,chen2021mvsnerf}, dynamic sampling realizes accurate depth-aware sampling by iteratively modifying the target sampling position according to the current position, \textit{i.e.}, in the sense of Markov Chain Monte Carlo (MCMC) \cite{MonteCarlo}. Leveraging the idea of MCMC, dynamic sampling progressively approaches the scene surface with just a few samples. To initialize, we randomly sample three points $s_{0}$, $s_{1}$ and $s_{2}$ in the interval, predict their density, and derive their accumulated transmittances, as shown in Fig.~\ref{sampling illu} (a)(1). Supposing that the distribution of the monotonically decreasing $T$ on the ray $\mathbf{r}(t)= \mathbf{o} + t\mathbf{d}$ is linear in a small sampling interval $D$,
we derive the distribution $\mathcal{U}_{1}$ of $T$ based on the $T(s_{1})$ and $T(s_{2})$. Then we sample the next point $s_{3}$ at the position where the expectation of $T$ is 0.5.
The iso-surface with $T$ equal to 0.5 is the implicit surface of NeRF and is usually used to obtain point clouds or meshes from NeRF. In this way, the sample is most likely to be the intersection of the surface and the ray. Then we predict the density of $s_3$, calculate $T(s_{3})$, and use it to update the distribution of the $T$ to $\mathcal{U}_{2}$ in the sampling interval. According to this rule, the position of the sampling point $x_{n}$ is decided by $t_{n}$ which satisfies:
\begin{equation}
      t_{n}=\left \{ t\mid E(T(s_{n}))=0.5, T\sim _{}\mathcal{U}_{n-1} \right \}.
\end{equation}
We repeat this \emph{Predict-then-Refine} operation cyclically until reaching the number of samples we set. To reduce the interference caused by background and depth estimation error, we also apply coarse uniform sampling before dynamic sampling. Note that DADS introduces no extra overhead since the total number of summations in progressive sampling is approximately equal to the number of summations in traditional NeRF with the same sample points.

With the cooperation of depth-aware sampling interval and dynamic sampling strategy, we find that the points approach the real surface very closely with a limited amount of computation, which means that our model is also capable of performing depth estimation in an unsupervised and generalizable way, in addition to the novel view rendering task. In the inference stage, we estimate a fine depth map from the neural radiance field by weighting and summing the depth values of the sampled points on marched rays (as is done in \cite{meng2021gnerf}). The most related work in depth estimation is NerfingMVS \cite{nerfing_2021_ICCV}, which is not generalizable. As for generalizable NeRF methods \cite{yu2021pixelnerf,wang2021ibrnet}, they have no access to accurate depth maps, diluting the geometric information during generalization for the pixel-scale optimization goal, as shown in Fig.~\ref{depthmap}. In our DARF, we introduce depth prior for every scene by DADS scheme, which helps DARF learn to infer the geometric layout of the scene and further achieve better rendering results.

\begin{figure*}[ht]
\centering
\includegraphics[width=1\linewidth]{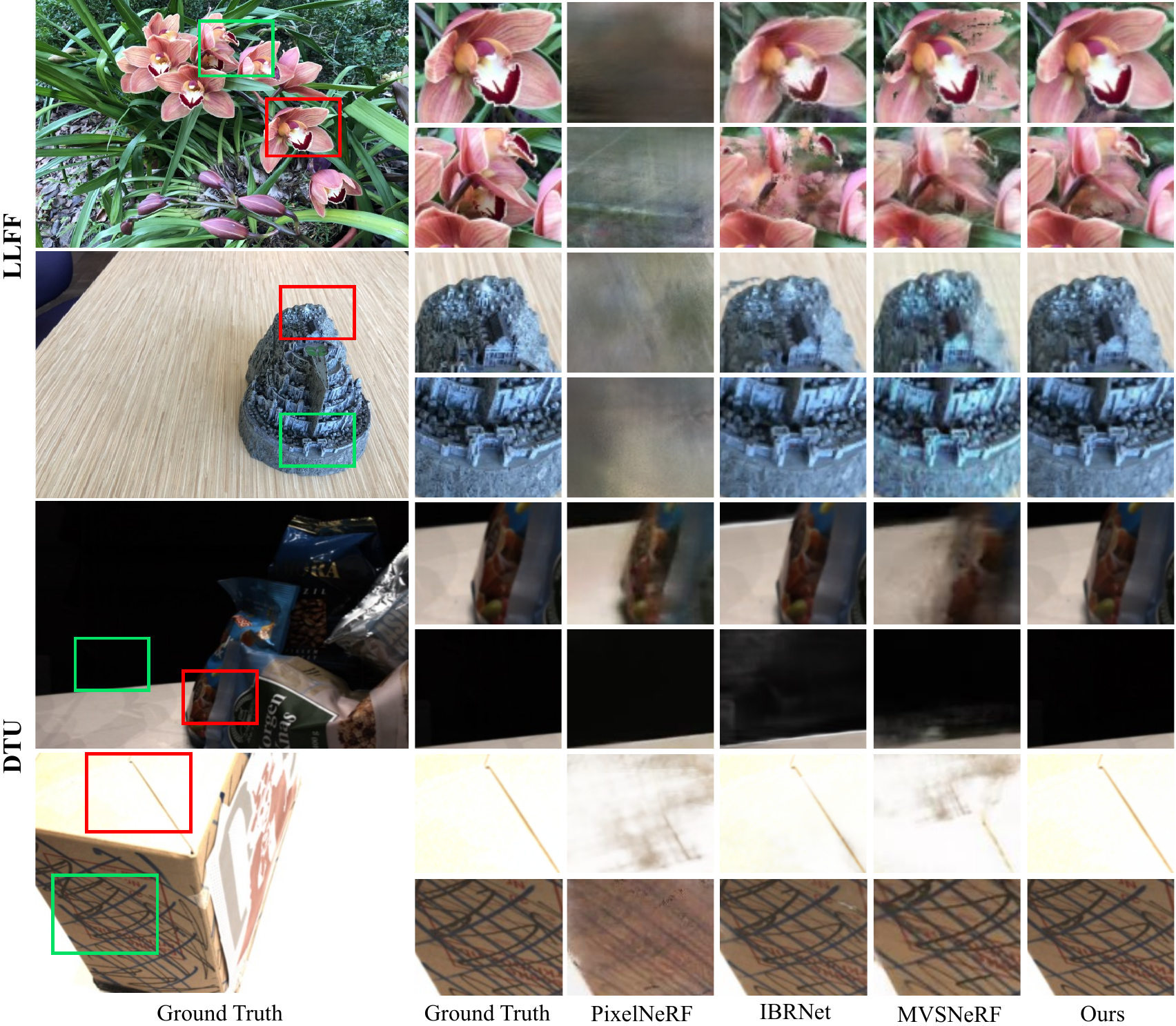}
\caption{Rendering quality comparison. For every scene, the contents in the red and green boxes are displayed in the first and second lines respectively. The rendering results of PixelNeRF are blurry. IBRNet has some deletions in details and marginal areas. MVSNeRF is not realistic in color and is blurred on geometric edges. Our results show high-quality rendering quality. The upper rows of zoom-in images correspond to contents in red boxes. }
\label{renderresult}
\end{figure*}

\subsection{Optimization Objectives}
We train the DARF framework based on the pixel value similarity between the rendered novel view image and its corresponding ground truth. In most NeRF methods, the rendering error is evaluated using only the squared error loss between the rendered and true pixel colors:
\begin{equation}
     \mathcal{L}_{s}=\sum_{r\in\mathcal{R}}^{}\left\|\hat{C_{r}}-C_{r}\right\|_{2}^{2},
\end{equation}
where $\mathcal{R}$ is the set of rays in each batch. 

\textbf{Multi-level Semantic Consistency Loss.} The above $\mathcal{L}_{s}$ can only constrain the similarity on pixel level and does not measure the overall semantic similarity of images. Thus, in order to learn more informative representation, we take the image content and the scene structure into account and propose a Multi-level Semantic Consistency Loss (MSC):
\begin{equation}
     \mathcal{L}_{\rm MSC}=\sum_{i=1}^{m}\left\|{\rm EL_{i}}({\rm S}(\hat{I}))-{\rm EL_{i}}({\rm S}(I))\right\|_{2}^{2},
\end{equation}
where $\hat{I}$ and $I$ are the predicted image and the ground truth image respectively. ${\rm EL(\cdot)_{i}}$ is an encoder, where $i$ indicates the index of the output layer of different depths. $m$ is the number of semantic scales, capturing multi-level semantic information. We utilize random sampling operation ${\rm S}(\cdot)$ and the Depth-Aware Dynamic Sampling Scheme (DADS) mentioned above to reduce computing costs. In practice, we take VGG-16 as the encoder and set $m$ equal to 3 empirically in our experiments. $\mathcal{L}_{\rm MSC}$ makes the network learn a semantic-level representation of scenes and take semantic coherence as contextual information to render more realistic novel-view images. 
Finally, the optimization objective of the DARF network is summarized as:
\begin{equation}
     \mathcal{L}=\mathcal{L}_{s}+\beta\mathcal{L}_{\rm MSC},
\end{equation}
where $\beta$ is the coefficient of MSC loss. 


\section{Experiments}
\subsection{Settings}
\noindent
\textbf{Datasets.}
We conduct experiments on DTU MVS Dataset (DTU) \cite{dtu} and Local Lightfield Fusion Dataset (LLFF) \cite{llff}, containing both indoor and outdoor scenes. We use camera parameters provided by the two datasets. In our experiment, we down-sample DTU to a resolution of $300\times400$ for training and evaluation. Images of LLFF are down-sampled to $640\times960$. For dividing training and test sets, we follow the partition of IBRNet \cite{wang2021ibrnet}. There is no intersection between training and test sets, ensuring the validity of generalization tests. For novel view rendering, we select five views as the input of the network and keep the sparsity of input views the same as \cite{wang2021ibrnet, yu2021pixelnerf}.

\noindent
\textbf{Evaluation Metrics.}
We evaluate image quality by PSNR, SSIM, and LPIPS \cite{lpips_2018_CVPR} which better reflects human perception. To explicitly demonstrate that DARF is able to reconstruct the scene's geometry better, \textit{e.g.}, has better capability on depth estimation, we derive depth maps and utilize the metrics Abs Rel, Sq Rel, RMSE and $\delta$ in \cite{nerfing_2021_ICCV} to evaluate the accuracy of depth maps quantitatively.

\noindent
\textbf{Implementation Details.} For the image encoder, we extract features by U-ResNet34. For the sample point feature decoder, we adopt the ray transformer structure with four heads from \cite{wang2021ibrnet}. We train every model using the training scenes and test it on unseen test scenes. We use Adam \cite{kingma2017adam} optimizer with an initial learning rate of $5\times {10}^{-4}$, which decays exponentially along with the optimization. We train our network using two RTX 2080 Ti GPUs, spending 30 hours with 64 samples and 10 hours with 32 samples. The hyper-parameter mentioned in the loss is set as $\beta=1$. 

\subsection{Results}
\label{results}
We compare with the state-of-the-art methods, including PixelNeRF \cite{yu2021pixelnerf}, IBRNet \cite{wang2021ibrnet} and MVSNeRF \cite{chen2021mvsnerf} in four aspects: (\textit{i}) The visual quality of novel-view rendering; (\textit{ii}) the geometry learning ability measured by depth estimation accuracy; (\textit{iii}) the computational efficiency of each method; and (\textit{iv}) intuitive evaluation of rendering results from users. All of the results are inferred without per-scene fine-tuning.

\renewcommand{\arraystretch}{1.7} 
\begin{table}[!htb]
\Large 
\caption{Quantitative comparison of rendering. The evaluation metrics are PSNR (higher is better), SSIM (higher is better) and LPIPS (lower is better). The best and second-best results are bold and underlined.}
\label{table1}
\resizebox{1\columnwidth}{!}{ 
\begin{tabular}{lcccccccccc}
\hline
\multirow{2}{*}{Method}&\multicolumn{3}{c}{DTU}&&\multicolumn{3}{c}{LLFF}\\\cline{2-4}\cline{6-8}
&PSNR$\uparrow$&SSIM$\uparrow$&LPIPS$\downarrow$&&PSNR$\uparrow$&SSIM$\uparrow$&LPIPS$\downarrow$\\\hline
PixelNeRF  & $20.62$ & $0.752$ & $0.392$ && $12.31$ & $0.416$ & $0.715$  \\
IBRNet   & $\underline{28.71}$ & $\underline{0.946}$ & $\underline{0.095}$ && $\underline{23.31}$ & $\underline{0.783}$ & $\underline{0.249}$  \\
MVSNeRF   & $26.84$ & $0.906$ & $0.195$ && $21.74$ & $0.764$ & $0.261$ \\\hline
\textbf{Ours} &	 \textbf{33.28}&
\textbf{0.971} & \textbf{0.086}&&   \textbf{24.10} & \textbf{0.845}& \textbf{0.218}\\
\hline
\end{tabular}}
\end{table}

\textbf{Novel-view rendering.}
We qualitatively compare the performance of our model with state-of-the-art methods in Fig.~\ref{renderresult}. We recover fine details more accurately and achieve the most realistic results on both the geometry level and appearance level. For simple scenes like the box, we achieve sharper textures, such as the lines on the first row of images. For complicated scenes like orchids, IBRNet produces blurred edges and textures, while MVSNeRF shows breaks on the petals. These issues likely arise from the model's inability to capture sufficient 3D information due to limited input. Additionally, the large sampling range and insufficient multi-view constraints lead to geometric errors, which result in inaccurate renderings from new perspectives. In contrast, our model benefits from a depth prior and the experience gained from pre-training data. We achieve more accurate geometry and appearance reconstruction by focusing sampling near the surface. Consequently, our results exhibit sharp edges and continuous, high-quality textures. 

The quantitative results are presented in Tab.~\ref{table1}. Our proposed model outperforms all competing methods across each evaluation metric and dataset. On the DTU dataset, we observe significant improvements, surpassing previous state-of-the-art methods by over 15.9\% in Peak Signal-to-Noise Ratio (PSNR). Additionally, our method achieves a 2.6\% gain in Structural Similarity Index (SSIM) and a 9.5\% improvement in Learned Perceptual Image Patch Similarity (LPIPS), further highlighting the effectiveness of our approach. Despite the more complex scenes in the LLFF dataset, and the smaller size of the LLFF dataset, our method still shows a notable improvement of 0.79 dB in PSNR, a 7.9\% increase in SSIM, and a 12.4\% improvement in LPIPS compared to the best-performing method. These results validate that by narrowing the sampling space and utilizing depth-aware sampling techniques, DARF excels in realistic novel-view rendering while maintaining strong generalization to unseen scenes. Our method effectively balances high-quality rendering and robustness across diverse scene complexities.

\begin{figure*}[t]
\centering
\includegraphics[width=1\textwidth]{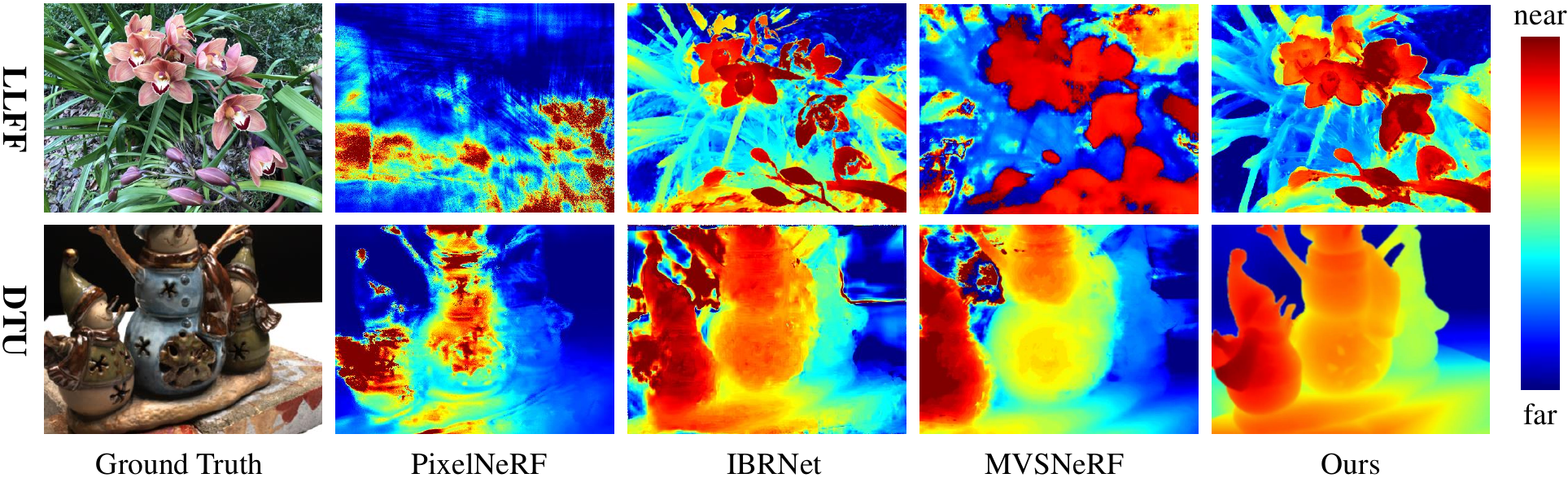}
\caption{Depth maps derived from four methods on DTU. Our method achieves significantly more accurate depth than the others, illustrating the effectiveness of the DADS. Our depth estimation results show clear boundaries and improved surface
continuity. The results also explain the advantage of our method in rendering new perspectives.}
\label{depthmap}
\end{figure*}
\textbf{Depth Estimation.}
DARF demonstrates a better awareness of scene geometry, which is proved by its better depth estimation, as shown in Fig.~\ref{depthmap}. Specifically, on the LLFF dataset, PixelNeRF performs poorly on depth estimation, which can be attributed to the limited training scenes in LLFF and PixelNeRF’s insufficient geometry learning capabilities. Although IBRNet performs better than PixelNeRF, it exhibits artifacts, such as breaks on the orchid petals. On the DTU dataset, all three comparison methods fail to distinguish between the background and foreground, due to the inherent geometric limitations of NeRF, which arise from full-space sampling and rendering-oriented optimization. In contrast, our depth estimation results show clearer foreground-background separation, sharper edges, and improved surface continuity (e.g., the visible geometric undulations on the snowman’s surface in the depth map). These improvements are attributed to the use of a depth estimation prior and a dynamic sampling strategy, which enable more accurate sampling and enhanced reconstruction of both the scene's geometry and appearance.

We use the metrics in \cite{nerfing_2021_ICCV,teed2018deepv2d} to evaluate the accuracy of depth estimation in Tab.~\ref{depthmetric}. The results are tested using the depth ground truth in DTU. We are the best in all metrics, showing an advantage in depth estimation. Our advantage on depth estimation illustrates that DARF is capable of reconstructing unseen scenes on both geometry and appearance.

\begin{table}[t]
\centering
\caption{Quantitative comparison of the depth estimation. The metrics are accuracy calculated by $l_{2}$ distance on the DTU dataset. The best and second-best results are bold and underlined.}
\resizebox{.47\textwidth}{!}{
\setlength\tabcolsep{1.2pt}{
\begin{tabular}{c|cccccc}
\hline
Method  &  Abs Rel$\downarrow$ & Sq Rel$\downarrow$ & RMSE$\downarrow$ & $\delta{<}1.25{\uparrow}$ & $\delta{<}1.25^{2}{\uparrow}$ & $\delta{<}1.25^{3}{\uparrow}$ \\

\hline
PixelNeRF &  $0.3943$ & $0.3127$ & $0.5329$ & $0.3673$ & $0.6212$ & $0.7843$  \\
IBRNet  &  $\underline{0.0712}$ & $\underline{0.0108}$ & $\underline{0.0961}$ & $\underline{0.9457}$ & $\underline{0.9965}$ & $\underline{0.9996}$  \\

MVSNeRF  &  $0.0986$ & $0.0239$ & $0.1243$ & $0.8691$ & $0.9547$ & $0.9812$ \\

Ours  &  $\textbf{0.0516}$ & $\textbf{0.0092}$ & $\textbf{0.0815}$ & $\textbf{0.9534}$ & $\textbf{0.9991}$ & $\textbf{0.9999}$ \\
\hline
\end{tabular}}}
\label{depthmetric} 
\end{table}

\renewcommand{\arraystretch}{1.5} 
\begin{table*}[ht]
\tiny 
\caption{Efficiency comparison on the number of sampling points. We compare the efficiency of different methods by sampling points in two sampling stages. Our method reduces samples by at least 25\% with higher PSNR. The best and second-best results are bold and underlined.}\label{efficiency}
\resizebox{2\columnwidth}{!}{ 
\begin{tabular}{lcccccccccccccc}
\hline
\multirow{2}{*}{Method} & \multicolumn{4}{c}{DTU} && \multicolumn{4}{c}{LLFF} \\ \cline{2-5} \cline{7-10}
& Samples & PSNR$\uparrow$ & SSIM$\uparrow$ & LPIPS$\downarrow$ && Samples & PSNR$\uparrow$ & SSIM$\uparrow$ & LPIPS$\downarrow$ \\ \hline
PixelNeRF & $128+64$ & $20.62$ & $0.752$ & $0.392$ & & $128+64$ & $12.31$ & $0.416$ & $0.715$ \\
IBRNet & $\underline{64+64}$ & $\underline{28.71}$ & $\underline{0.946}$ & $\underline{0.095}$ & & $\underline{64+64}$ & $\underline{23.31}$ & $\underline{0.783}$ & $\underline{0.249}$ \\
MVSNet & $128$ & $26.84$ & $0.906$ & $0.195$ & & $128$ & $21.74$ & $0.764$ & $0.261$ \\\hline
\textbf{Ours} 
& $\mathbf{32+32}$ & $\mathbf{29.15}$ & $\mathbf{0.952}$ & $\mathbf{0.091}$ &
& $\mathbf{32+32}$ & $\mathbf{23.61}$ & $\mathbf{0.801}$ & $\mathbf{0.235}$ \\
\hline
\end{tabular}}
\end{table*}

\begin{table}[h]
\caption{Efficiency comparison on the rendering time (s) and FLOPs (TMac). The results are tested during one inference on the DTU dataset, with the resolution of $300\times400$.}
\centering
\tiny
\resizebox{.48\textwidth}{!}{
\setlength\tabcolsep{1pt}{
\begin{tabular}{l| c c c c}
\hline
 & \multicolumn{1}{c}{PixelNeRF} & \multicolumn{1}{c}{IBRNet} & \multicolumn{1}{c}{MVSNeRF} & \multicolumn{1}{c}{Ours}\\
\hline
Samples$\downarrow$  & $128+64$ & $64+64$ & $128$ & $\textbf{32+32}$ \\
Time (s)$\downarrow$  & $14.3$ & $11.1$ & $9.8$ & $\textbf{7.2}$ \\
FLOPs (TMac)$\downarrow$  & 4.43 & 3.48 & 3.25 & \textbf{2.31} \\
\hline
\end{tabular}}}
\label{tab:flop}  
\end{table}


\textbf{Efficiency.} First, we compare the efficiency based on the number of sampling points, as shown in Tab.~\ref{efficiency}. The number of samples is the sum of sample points in two sampling stages. On both two datasets, we achieve better PSNR than the state-of-the-art methods when the number of sampling points is reduced by 50\%. Compared with other methods, our approach samples fewer points on the ray, which reduces the time of the forward pass, calculation, and summation of volume rendering during training and inference.

To be more intuitive, we also provide inference times and FLOPs in Tab.~\ref{tab:flop}, with settings the same as Tab. \ref{table1}. The results are tested on the DTU dataset and the resolution of the image is $300\times400$. Inference time is affected by the device and related to the equipment, which can only be used as a rough indicator for efficiency evaluation. Compared with other methods, we reduce the inference time by 26.5\% than the second-fastest method MVSNeRF while making a 23.9\% improvement on the PSNR. Compared with the second-best method IBRNet, the efficiency improvement is 35.1\% on the inference time and 33.6\% on the FLOPs while achieving better rendering quality. When we use the same samples with IBRNet and MVSNeRF, the PSNR is 33.28dB as illustrated in Tab. \ref{table1}, which far outperforms other methods.

\begin{figure}[ht]
\includegraphics[width=1\linewidth]{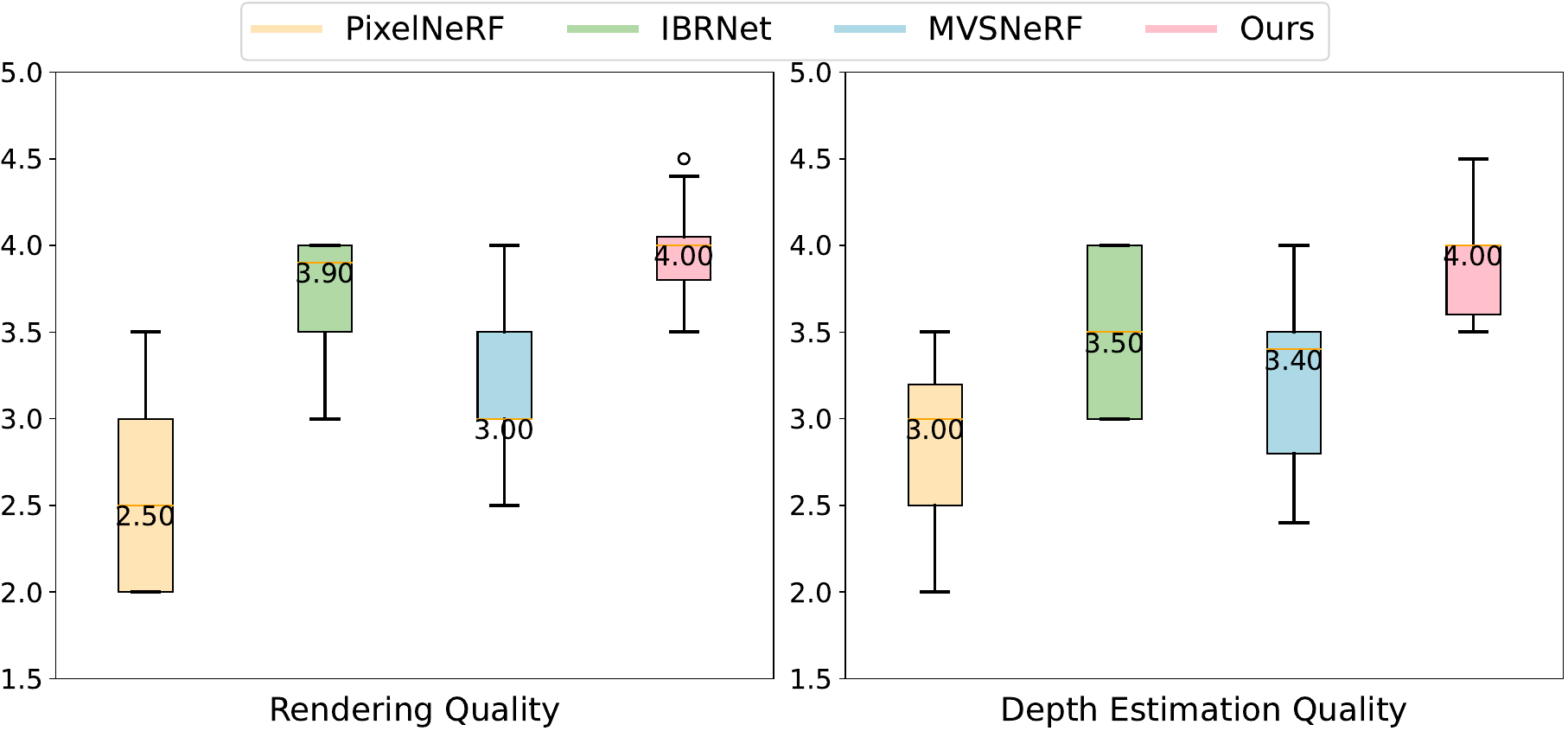}
\caption{Boxplot illustration of user study. Our method
demonstrates better performance (high means) and stability
across various test scenes (narrow interquartile range).}
\label{fig: user}
\end{figure}

\textbf{User Study.} To ensure a comprehensive evaluation, we introduce a user study to further assess the performance of our method. We conducted experiments with 20 participants who evaluated the results of 5 randomly selected scenes. For each scene, participants were provided with 4 images rendered from novel viewpoints, accompanied by the corresponding depth maps. Each participant was asked to assess 10 sets of results from anonymous methods and assign a preference score (ranging from 1 to 5) for both rendering quality and depth estimation accuracy. As illustrated in Fig. \ref{fig: user}, we present the distribution of scores, including medians, means, quartiles, and outliers. Our results indicate that our method is significantly preferred over all baseline methods in terms of both rendering quality (median = 4.0) and depth accuracy (median = 4.0). Moreover, the narrow interquartile range of our method highlights its more consistent and robust performance across various scenes, further demonstrating its reliability in different contexts.

\subsection{Ablation Study}
\begin{figure}[h]
\includegraphics[width=1.0\linewidth]{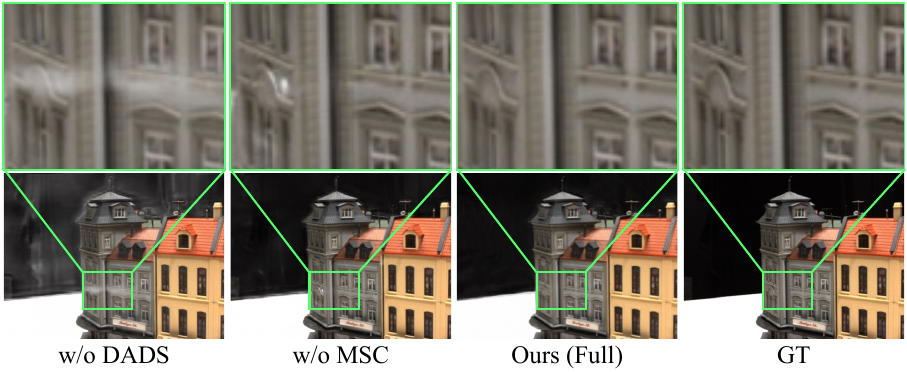}
\caption{Ablation Study (qualitative results): We remove components from our full model and showcase qualitative results, demonstrating the visual impact of each component.}
\label{fig:abla}
\end{figure}

\textbf{Component Analysis.} We design three variants based on our full model to validate the effectiveness of each component in our model. We present qualitative and quantitative results on the DTU dataset in Fig. \ref{fig:abla} and Tab. \ref{tab:abla}, respectively. The `w/o' stands for `without'.

\begin{table}[t]
\centering
\caption{Ablation study (quantitative results). We remove components from our full model and present quantitative results on rendering quality and depth estimation. The best results are bold.}
\label{tab:abla}
\resizebox{0.47\textwidth}{!}{
\begin{tabular}{c|ccc|ccc}
\toprule 
Tasks &   \multicolumn{3}{c|}{View Synthesis} & \multicolumn{3}{c}{Depth Estimation} \\
\midrule
Metrics & PSNR$\uparrow$ & SSIM$\uparrow$ & LPIPS$\downarrow$ & Abs Rel$\downarrow$ & Sq Rel$\downarrow$ & $\delta{<}1.25{\uparrow}$ \\
\midrule
w/o DADS & 29.13 & 0.948 & 0.093 & 0.0556 & 0.0101 & 0.9499 \\
w/o MSC & 32.87 & 0.960 & 0.088 & 0.0519 & 0.0094 & 0.9532 \\
\textbf{Ours (Full)} & \textbf{33.28} & \textbf{0.971} & \textbf{0.086} & \textbf{0.0516} & \textbf{0.0092} & \textbf{0.9534} \\
\bottomrule 
\end{tabular}}
\end{table}

We first examine the benefits of the DADS in our framework. As shown in Table~\ref{tab:abla} and the visual results in Figure~\ref{fig:abla}, DADS significantly enhances both rendering quality and depth estimation accuracy, achieving a remarkable 4.15dB improvement in PSNR and a 7.2\% reduction in Abs Rel. In terms of visualization, DADS mitigates issues such as broken textures and inconsistencies in image appearance, particularly in areas with thin lines and edges, improving the realism of colors and overall coherence.

MSC is particularly impactful for enhancing rendering quality. For view synthesis, MSC achieves a 0.41dB improvement in PSNR, helping the model learn more informative representations and produce results with high fidelity. This also echoes the result, MSC makes the rendered details closer to the ground truth.


\begin{figure*}[t]
\centering
\includegraphics[width=1\textwidth]{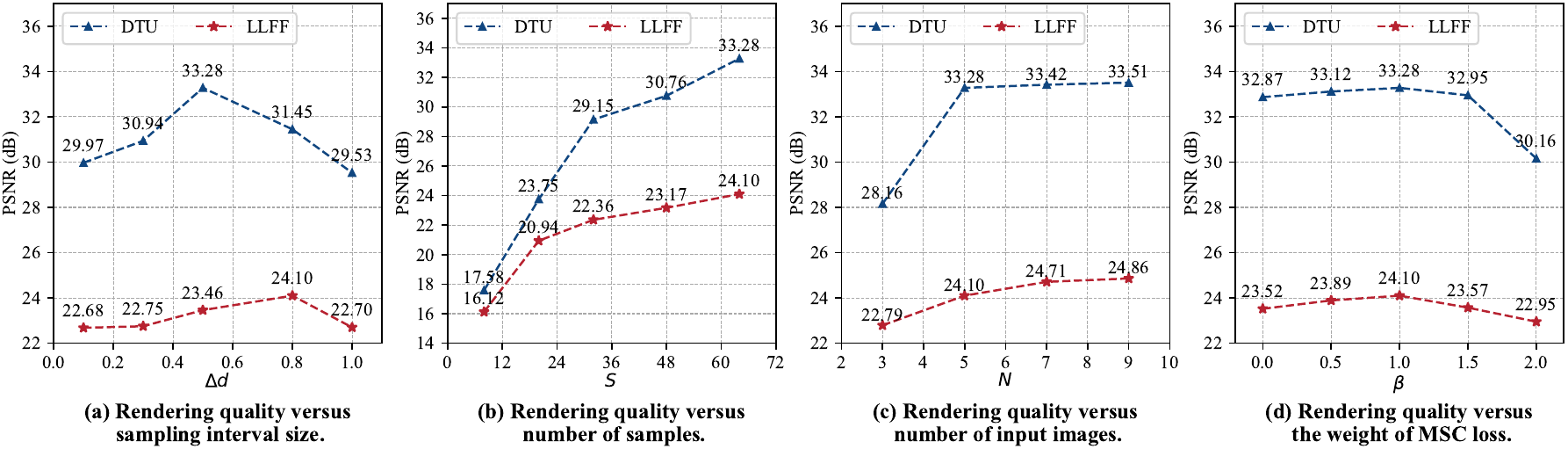}
\caption{Results of parameter analysis. We show the change in rendering quality, with the size of the sampling interval (a), the decrease of samples (b), the number of input images (c), and the weight of MSC loss (d).}
\label{fig_ablation}
\end{figure*}

\textbf{Parameter Analysis.} Three important parameters introduced by our method are analyzed in this section. The first is the size of the sampling interval $\Delta d$, which is used to control the sampling range. As Fig.~\ref{fig_ablation} (a) shows, we derive the best results at $\Delta d=0.5$ on DTU and $\Delta d=0.8$ on LLFF. As $\Delta d$ decreases, the quality of rendered images decreases, as there is less opportunity to fix depth estimation errors. When the sample interval $\Delta d$ increases, the awareness of geometry decreases, leading to worse results.
The second parameter is the number of sampling points. We show the results in Fig.~\ref{fig_ablation} (b). Before the sampling points of each stage drop to 20, image quality is comparable to other methods with 128 sample points, demonstrating that DARF has a good rendering capability even when the sampling points are reduced. However, the image quality is noticeably degraded when the samples are further reduced. The reason is that the coarse depth estimation is not absolutely accurate and needs to be further corrected during sampling. 
Another parameter is the number of input views. The analysis results are shown in Fig.~\ref{fig_ablation} (c). With increasing the number of input images, the quality of the novel view rendering is gradually improved, enabling high-quality visual results. The last parameter is the weight $\beta$ of the MSC loss. In Fig.~\ref{fig_ablation} (d), as the weight of MSC increases, the PSNR initially increases, reaches its peak when $\beta$ equals 1, and then starts to decrease. In the early stages, MSC helps to preserve perceptually meaningful structures, textures, and patterns through the feature-level similarity constraint. However, as the weight increases further, MSC interferes with the primary role of the reconstruction loss $\mathcal{L}_s$, leading to a negative impact. Therefore, we set $\beta=1$ to achieve the optimal overall performance.

\begin{figure}[h]
\centering
\includegraphics[width=.95\linewidth]{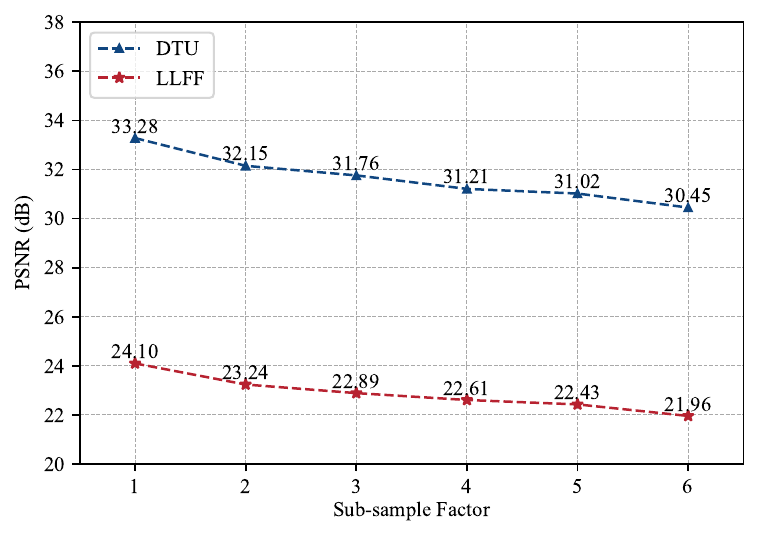}
\caption{Sensitivity to source view density. We subsample the views by variant factors to create varying source view densities. The performance degradation is in a reasonable range.}
\label{fig_density}
\end{figure}

\subsection{Robustness Analysis}
We investigate the robustness of our approach to the sparsity of the source views. For a target view, we sample $n$ input images from sets containing different numbers of images following \cite{wang2021ibrnet}, each of which contains $n$, $2n$, $3n$, $4n$, $\cdots$ images closest to the target view. We define the sub-sample factor as the ratio of the sampling set size to the number of input source images, i.e., the coefficient before $n$. To investigate the robustness of DARF, we analyze its sensitivity to source view density by changing the sub-sample factor. The sub-sample factor is larger when the view density is sparser, i.e., the angles between the source views are larger. We conduct experiments on different sub-sample factors while keeping other settings the same as the Tab. \ref{table1} of the paper. The results are shown in Fig.~\ref{fig_density}. 

With the sub-sample factor increasing, i.e., the view density reducing, DARF achieves the best rendering quality at $1$ and slowly degrades. When the density reduces to one-sixth of the initial, the results on DTU are still better than the state-of-the-art methods in Tab. \ref{table1}. The results prove that DARF is robust to the large angle of source images thanks to its depth-aware ability.

\subsection{Failure Cases}
\begin{figure}[h]
\centering
\includegraphics[width=.95\linewidth]{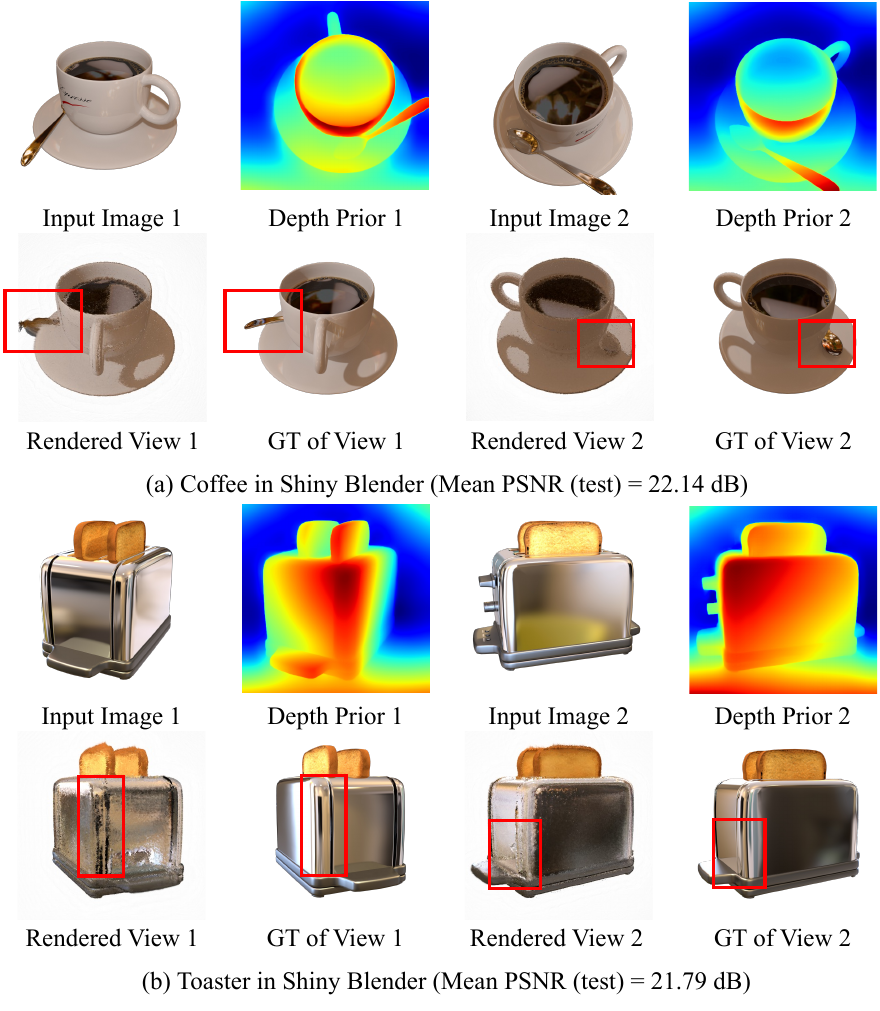}
\caption{Failure cases: (a) Coffee, with reflective water surface and spoon. (b) Toaster, which is made of reflective metal and in a light environment. These reflective surfaces introduce ambiguities in the depth map, as reflections disrupt the model's ability to distinguish multiple objects, leading to inaccurate depth predictions.}
\label{fig: rebu_failer}
\end{figure}

Since we rely on a pre-trained depth estimation model (Depth Anything v2) to obtain the depth prior, some incorrect depth estimations are inevitable. To provide a more comprehensive evaluation of DARF's performance and offer valuable insights for future work, we test our model on more challenging scenes. Specifically, we choose the Shiny Blender dataset \cite{verbin2022ref}, which features a variety of materials and prominent reflections. Two failure cases are shown in Fig.~\ref{fig: rebu_failer}. In these cases, Depth Anything v2 struggles with reflective objects and low-texture areas. Consequently, the performance of DARF is influenced by the depth prior, resulting in a loss of high-frequency details in both scenarios. Despite the limitations of the depth prior in these cases, DARF demonstrates a degree of correction ability and ultimately provides reasonably acceptable rendering results.

To further improve the quality of these challenging cases, we
propose two potential solutions: (\textit{i}) Use a more powerful
prior model in the depth estimation module, such as a depth estimation model that is robust to reflections or a multi-view consistent depth estimation model. (\textit{ii}) Incorporate real sparse depth priors as supervision. This approach would assist our model in better estimating the sampling intervals, thereby maintaining both quality and efficiency even in such extreme conditions.

\section{Conclusions}
In this paper, we introduce the Depth-Aware Generalizable Neural Radiance Field (DARF), which incorporates a Depth-Aware Dynamic Sampling (DADS) strategy and a Multi-Level Semantic Consistency (MSC) constraint. By leveraging depth priors from a foundation model trained on a large-scale dataset, we focus the sampling in regions with the highest probability of surface distribution. The DADS strategy dynamically adjusts the distribution of sampling points, concentrating them near the surface. This allows us to achieve high-quality rendering with fewer sampling points. Moreover, the depth guidance enhances the model's robustness to the sparsity of source views. Extensive experiments demonstrate that DARF is capable of synthesizing realistic free-viewpoint images and generating accurate depth maps, while significantly reducing training and inference time. In future work, we aim to explore techniques for learning additional geometric information to more accurately render and reconstruct scenes and objects.

\section*{Declaration of competing interest}
The authors declare that they have no known competing financial interests or personal relationships that could have appeared to influence the work reported in this paper.

\section*{CRediT authorship contribution statement}
\textbf{Yue Shi}: Conceptualization, Methodology, Software, Writing - Original draft preparation. \textbf{Dingyi Rong}: Software, Validation. \textbf{Chang Chen}: Validation, Writing - Original draft preparation. \textbf{Chaofan Ma}: Visualization, Validation. \textbf{Bingbing Ni}: Conceptualization, Supervision, Writing - Reviewing \& Editing. \textbf{Wenjun Zhang}: Supervision, Writing - Reviewing \& Editing.

\section*{Acknowledgement}
This work was supported by the National Natural Science Foundation of China (62431015).

\printcredits

\bibliographystyle{model1-displa-num-names}



\end{document}